\documentclass[letterpaper]{article} 
\usepackage{aaai25}  
\usepackage{times}  
\usepackage{helvet}  
\usepackage{courier}  
\usepackage[hyphens]{url}  
\usepackage{graphicx} 
\usepackage{natbib}
\usepackage{caption} 
\usepackage{algorithm}
\usepackage{algorithmic}
\usepackage{amsfonts}
\usepackage{amsmath}
\usepackage{multicol}
\usepackage{multirow}
\usepackage{adjustbox}
\usepackage{xcolor,colortbl}
\usepackage{booktabs}
\usepackage{xcolor}
\usepackage{array}
\usepackage[utf8]{inputenc}
\DeclareMathOperator*{\softmax}{\mathtt{softmax}}
\DeclareMathOperator*{\argmax}{\mathtt{argmax}}
\DeclareMathOperator*{\argmin}{\mathtt{argmin}}

\DeclareMathOperator*{\maxa}{\mathtt{max}}

\usepackage{mathtools}
\usepackage{newfloat}
\usepackage{listings}
\DeclareCaptionStyle{ruled}{labelfont=normalfont,labelsep=colon,strut=off} 
\floatstyle{ruled}
\newfloat{listing}{tb}{lst}{}
\floatname{listing}{Listing}
\DeclareCaptionStyle{ruled}{labelfont=normalfont,labelsep=colon,strut=off} 
\floatstyle{ruled}
\newfloat{listing}{tb}{lst}{}
\floatname{listing}{Listing}
\title{

HiGDA: Hierarchical Graph of Nodes to Learn Local-to-Global Topology \\ for Semi-Supervised Domain Adaptation

}
\author{
    Ba Hung Ngo\textsuperscript{1}\equalcontrib,
    Doanh C. Bui\textsuperscript{2}\equalcontrib,
    Nhat-Tuong Do-Tran\textsuperscript{3},
    Tae Jong Choi\textsuperscript{1}\footnote{Corresponding author.}
}
\affiliations{
    \textsuperscript{\rm 1}Graduate School of Data Science, Chonnam National University, South Korea\\
    \textsuperscript{\rm 2}School of Electrical Engineering, Korea University, South Korea\\
    \textsuperscript{\rm 3}Department of Computer Science, National Yang Ming Chiao Tung University, Taiwan\\
    ngohung@chonnam.ac.kr, doanhbc@korea.ac.kr, tuongdotn.cs11@nycu.edu.tw, ctj17@jnu.ac.kr
}

\usepackage{bibentry}

\begin{document}

\maketitle

\begin{abstract}


The enhanced representational power and broad applicability of deep learning models have attracted significant interest from the research community in recent years. However, these models often struggle to perform effectively under domain shift conditions, where the training data (the \textit{source domain}) is related to but exhibits different distributions from the testing data (the \textit{target domain}). To address this challenge, previous studies have attempted to reduce the domain gap between source and target data by incorporating a few labeled target samples during training—a technique known as semi-supervised domain adaptation (SSDA). While this strategy has demonstrated notable improvements in classification performance, the network architectures used in these approaches primarily focus on exploiting the \textit{features} of individual images, leaving room for improvement in capturing rich representations. In this study, we introduce a Hierarchical Graph of Nodes designed to simultaneously present representations at both \textit{feature} and \textit{category} levels. At the \textit{feature level}, we introduce a local graph to identify the most relevant patches within an image, facilitating adaptability to defined main object representations. At the \textit{category level}, we employ a global graph to aggregate the features from samples within the same category, thereby enriching overall representations. Extensive experiments on widely used SSDA benchmark datasets, including \textit{Office-Home}, \textit{DomainNet}, and \textit{VisDA2017}, demonstrate that both quantitative and qualitative results substantiate the effectiveness of HiGDA, establishing it as a new state-of-the-art method.

\end{abstract}

\section{Introduction}

Over the past decade, deep learning models have achieved remarkable success in various vision-based tasks, largely based on the assumption that the training and testing datasets share identical distributions. However, in real-world applications, the training samples (source domain) are often collected from a domain that differs from the testing samples (target domain). As a result, models trained on the source domain frequently underperform when applied to the target domain due to the domain shift problem \citep{domainshift}. To address this issue, domain adaptation (DA) methods \cite{DANN,MCD,ECACL,MCL,SLA} have been proposed to bridge the domain gap between source and target domains at both the \textit{feature} and \textit{category} levels.

\begin{figure} 
\centerline{\includegraphics[width=0.48\textwidth]{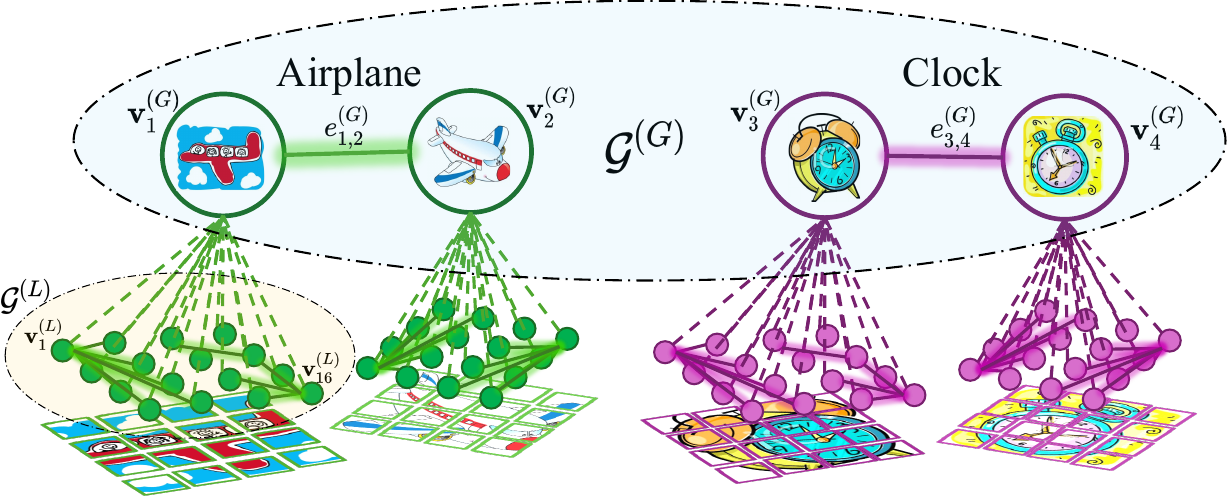}}
\caption{Overview of the hierarchical graph of nodes. Input images are considered global nodes, each comprising multiple sub-patches that are considered local nodes.}
\label{fig:overview} \vspace{-1.6em}
\end{figure}

At the \textit{feature level}, early DA approaches \citep{DANN,MCD} employed convolutional neural networks (CNNs) as the standard backbone to encode the semantic representations of input images. CNNs \citep{CNN} use sliding convolutions with learnable kernels to introduce shift-invariance. As a result, the CNN architecture tends to preserve local spatial relationships, extracting discriminative features that cover only a specific region of pixels at a time. Consequently, the CNN backbone is not flexible enough to capture complex objects, such as images consisting of multiple similar contexts located at different positions. In contrast, recent DA methods \citep{pathmix, tvt} leverage the cross-attention mechanism in the vision transformer (ViT) \citep{ViT} to better comprehend image features. The ViT architecture treats an image as a sequence of patches and explores their relationships, effectively capturing global relations at the \textit{feature level}. However, it struggles to recognize representations of the main object when the input images contain irrelevant objects or a dominant background, as it assumes all patches in an image are correlated.
To address this problem, \citep{Ngo_2024_CVPR} introduces a hybrid approach that leverages the advantages of both ViT and CNN models. These models complement each other's information via a co-training strategy, alleviating their weaknesses and thereby enhancing image classification performance in CNN during inference. Although the hybrid model significantly improves classification performance, two concerns remain: 1) In the case of ViT, all patches are assumed to have the same level of correlation with each other, which allows various types of noise to be attended to, thus hindering comprehensive representation exploration. 2) As for the hybrid approach, it results in a large number of parameters that need to be optimized. Based on these observations, we pose the following question: \textit{\textbf{Q1: How can we explicitly present the semantic information of an image at the \textit{feature level} with a compact model?}}

At the \textit{category level}, alignment across source and target domains can be effectively achieved through pseudo-labeling, where generated pseudo-labels from the unlabeled target samples are used to update the parameters of the trained model. Additionally, several semi-supervised domain adaptation (SSDA) methods \citep{ProML,SLA,EFTL} have demonstrated that incorporating one or three labeled target samples per class during the training phase, alongside labeled source data, can substantially enhance classification accuracy. The primary objective of the pseudo-labeling process or the inclusion of a few labeled target samples is to increase the representation of the target domain in the learning model. These studies reveal that the explicit presentation of features is a crucial process in domain adaptation tasks. 
However, it is important to note that previous DA studies often overlook the structural information of labeled samples during training. As a result, samples within the same class may consist of similar patterns that are not effectively integrated to enrich features. Based on this observation, another question arises: \textit{\textbf{Q2: How can connections be established among samples belonging to the same category for feature aggregation to achieve better representations?}}

In this study, we propose a \textbf{Hi}erarchical \textbf{G}raph of nodes for \textbf{D}omain \textbf{A}daptation (HiGDA), as illustrated in Fig. \ref{fig:overview}, to address the two aforementioned questions. HiGDA is designed with two levels: the \textit{local graph} and the \textit{global graph}. The \textit{local graph} is constructed to exploit the \textit{feature level} of each individual image. Similar to ViT, we partition an image into patches and arrange these patches into a sequence, with each patch conceptualized as a \textit{local node}. However, unlike ViT, our local graph is constructed with adaptable connections, where local nodes establish pairwise connections solely with their closest neighbors for feature aggregation. In this approach, features within an image are leveraged through a graph-based methodology. By doing so, the impact of noise is mitigated since the representations of the main object are kept at a distance from irrelevant objects and background noise in the embedding space. The output features of the \textit{local graph} are then used to build the \textit{global graph}. Here, each image is conceptualized as a \textit{global node}. Next, we construct \textit{global graphs} to explore the category-level representations. Specifically, representations of \textit{global nodes} belonging to the same category are concatenated to
enrich the context of visual recognition, thereby addressing the concern raised in \textit{\textbf{Q2}}.

Furthermore, to enhance the alignment of representations across domains at the \textit{category level}, we propose a Graph Active Learning (GAL) algorithm. In GAL, the trained HiGDA model generates pseudo-labels from unlabeled target samples in each training episode. These pseudo-labels are then combined with given labeled samples to construct the \textit{global graphs}. As a result, the domain gap between the source and target domains at the \textit{category level} is minimized.

In summary, our contributions are as follows:

\begin{itemize}
\item We propose a hierarchical graph of nodes (HiGDA) to explore the local-to-global structure in representations associated at \textit{feature} and \textit{category} levels.

\item We demonstrate that HiGDA, structured as a graph, is more compact and efficient in model size than existing methods while achieving impressive results.

\item We introduce the GAL strategy, which efficiently uses pseudo-labels generated from unlabeled target samples to enhance cross-domain representations at the \textit{category level}.

\item We validate HiGDA on various domain adaptation datasets under the SSDA setting. Extensive experimental results demonstrate HiGDA’s effectiveness, achieving state-of-the-art performance.
\end{itemize}

\begin{figure*}[http]
\centerline{\includegraphics[width=0.85\textwidth]{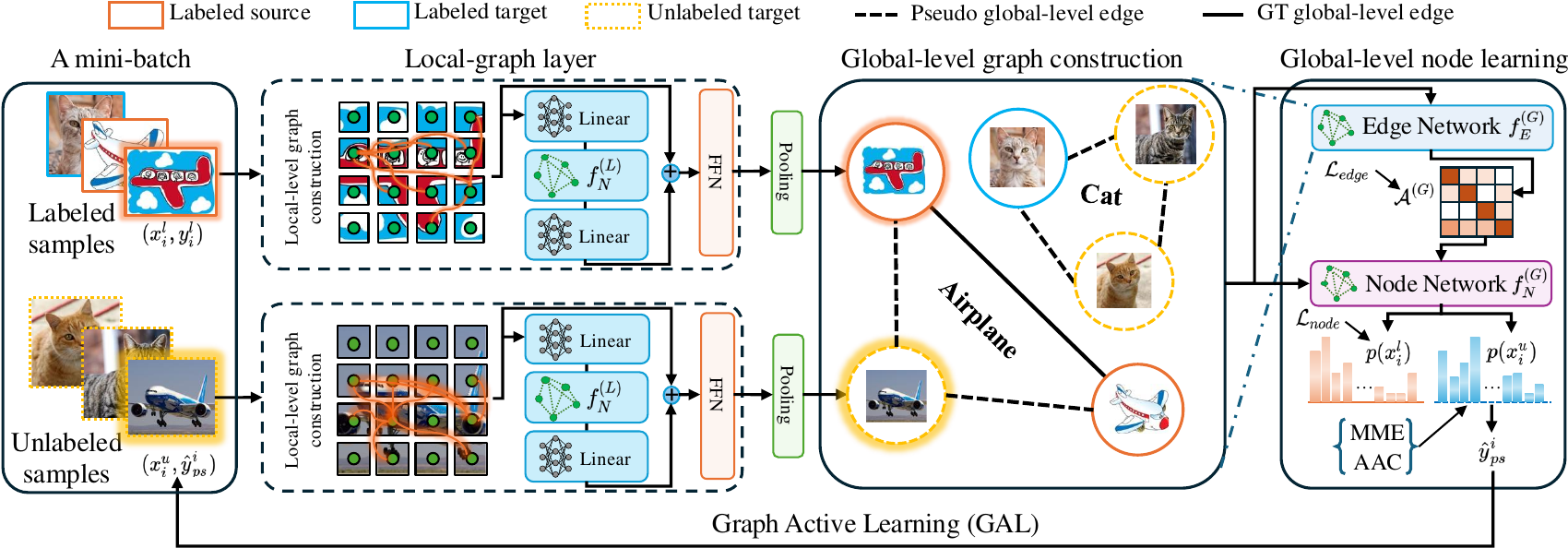}}
\caption{Overview of HiGDA. Each image forms a local graph $\mathcal{G}^{(L)}$, and a mini-batch forms a global graph $\mathcal{G}^{(G)}$ to explore feature to \textit{category level} representations. Then, GAL minimizes bias from the source dataset.}
\label{fig:higraph}
\end{figure*}

\section{Related Work}

\noindent \textbf{Network Architectures in SSDA}

\noindent Early SSDA methods \citep{MME, ECACL, APE, AAC} follow a similar methodology when designing network architectures, typically comprising two main components: convolutional neural networks (CNNs) as feature extractors and multilayer perceptrons (MLPs) as single classifiers. The various network architectures are evolving rapidly by flexibly incorporating multiple components to improve classification accuracy. The representative works include DECOTA \citep{DECOTA}, UODA \citep{UODA}, and ASDA \citep{ASDA}.  DECOTA introduces a decomposing framework consisting of two distinct branches that share the same network architecture (CNN+MLP). Each branch is trained on a different labeled dataset, and they exchange knowledge using a co-training strategy to achieve consistency on the unlabeled samples. Similarly, UODA and ASDA use a CNN feature extractor followed by two MLP classifiers to provide two different views. They demonstrate that the data bias in SSDA can be alleviated by using this type of network architecture. However, CNNs are only effective in capturing local spatial hierarchical representations. Therefore, using CNNs as the feature extractor faces challenges in achieving a comprehensive global visual context. Furthermore, these network architectures solely focus on exploiting the semantic representations of each individual image (\textit{feature-level}).

\noindent \textbf{Graph Neural Networks}

\noindent Various Graph Neural Network (GNN) models have been developed to address graph learning tasks on graph-structured data, such as biochemical, social, or citation datasets. A prominent example of GNNs is the Graph Convolutional Network (GCN) \citep{gcn}, which employs the adjacency matrix in conjunction with convolution operations. These graph-based models have consistently demonstrated impressive performance in numerous graph-learning tasks.  Consequently, they have gained widespread adoption in domain adaptation tasks \citep{gcan,TFLG,GABC,SPA}. For instance, \citep{gcan} employs the triplet loss to construct the adjacency matrix, incorporating similarity information of source and target samples. In contrast, \citep{TFLG} and \citep{GABC} build graphs using pairwise label similarity, establishing cross-domain connectivity if the ground-truth labels of source samples share the same information with pseudo labels generated from the unlabeled target samples. However, the aforementioned approaches primarily focus on constructing graphs at the higher level (\textit{category level}), neglecting the lower level (\textit{feature level}). Consequently, it becomes challenging to effectively present features at the \textit{category level}.

\vspace{-0.1cm}
\section{Methodology}
\vspace{-0.1cm}
\subsection{{Problem definition}} 

\noindent In semi-supervised domain adaptation (SSDA), we are given substantial labeled set from the source domain, $S = \{x^{s}_i, y_{i}^{s}\}^{N_{s}}_{i=1}$, alongside a much smaller labeled set from the target domain, $T = \{x_i^t, y_i^t\}^{N_t}_{i=1}$, to form a combined labeled set $D_l = S \cup T = \{x^l_i, y_i^l\}^{N_s + N_t}_{i=1}$, where $N_{s} \gg N_{t}$. The final goal is to minimize the classification error on the unlabeled target set, $D_u = \{x_i^u\}^{N_{u}}_{i=1}$. 

\vspace{-0.2cm}
\subsection{{Overview of hierarchical graph design}}
\label{sec:overview}

Figure \ref{fig:overview} illustrates our methodology in designing a hierarchical graph of nodes for SSDA. In this approach, we introduce a local-to-global topology to present a hierarchical graph of nodes, $\mathcal{G} = \{\mathcal{G}^{(L)}, \mathcal{G}^{(G)}\}$, where $\mathcal{G}^{(L)}$ and $\mathcal{G}^{(G)}$ are the local and global graphs, respectively, for representing features across domains. At the \textit{feature level}, an image is split into sub-patches, with each patch conceptualized as a \textit{local node}. Then, $\mathcal{G}^{(L)}$ is designed to explicitly produce the semantic representations of each image by aggregating the embeddings of local nodes with their neighbors. At the \textit{category level}, output features of each input image are considered as a \textit{global node}, which is handled by $\mathcal{G}^{(G)}$. Specifically, $\mathcal{G}^{(G)}$ is created to aggregate global nodes sharing to the same category.

\subsubsection{{Local graph.}}
Given an image sample $x_{i} \in \mathbb{R}^{H \times W \times 3}$, we conceptualize it as a local graph $\mathcal{G}^{(L)} = \{\mathcal{V}^{(L)}, \mathcal{E}^{(L)}\}$. In detail, we define a set of local nodes $\mathcal{V}^{(L)} = \{\mathbf{v}^{(L)}_i\}^{N^{(L)}}_{i=1}$, where $N^{(L)}$ is the number of local nodes, which is fixed for all samples, and each local node $\mathbf{v}^{(L)}_i$ represents a feature vector of a sub-patch image tiled from $x_{i}$. Regarding the construction of the local edge set $\mathcal{E}^{(L)}$, an edge $e^{(L)}_{i,j} \in \mathcal{E}^{(L)}$ connecting $\mathbf{v}^{(L)}_i$ and $\mathbf{v}^{(L)}_j$ exists if and only if $\mathbf{v}^{(L)}_j \in \mathcal{N}(\mathbf{v}^{(L)}_i)$, where $\mathcal{N}(\mathbf{v}^{(L)}_i)$ is the set of $k$-nearest neighbors of $\mathbf{v}^{(L)}_i$. 

\subsubsection{{Global graph.}}
Similarly the local graph, we construct a global graph, $\mathcal{G}^{(G)} = \{\mathcal{V}^{(G)}, \mathcal{E}^{(G)}\}$. A set of global nodes is defined as $\mathcal{V}^{(G)} = \{\mathbf{v}^{(G)}_i\}_{i=1}^{N^{(G)}}$, where each global node $\mathbf{v}^{(G)}_i$ is output feature vector of the image sample $x_i$ handled by the local graph. $N^{(G)}$ denotes the number of global nodes. To determine the edge set $\mathcal{E}^{(G)}$, an edge $e^{(G)}_{i,j} \in \mathcal{E}^{(G)}$ between two global nodes exists if and only if they have the same categories, \textit{i.e.}, $y_{i}$ = $y_{j}$.

\vspace{-0.2cm}
\subsection{{HiGDA learning process}}
In this section, we present the HiGDA learning process including local-graph ($LoG$) and global-graph ($GoG$) networks, as illustrated in Fig. \ref{fig:higraph}.
\vspace{-0.2cm}
\subsubsection{{Local-graph network.}}

To learn the local graph \(\mathcal{G}^{(L)}\), we design a local-graph network \( {LoG} \). This network incorporates an embedding network for image patching to construct nodes and a series of local-graph layers. The core of these layers is the local-node network \( f^{(L)}_N \), which is primarily responsible for performing message passing, i.e., exchanging information between nodes.

Given $x_{i} \in \mathbb{R}^{H \times W \times 3}$, it is divided into sub-patches and their corresponding embeddings as a set of local nodes $\mathcal{V}^{(L)} = \{\mathbf{v}^{(L)}_i\}_{i=1}^{N^{(L)}}$ by the embedding network. Each local node $\mathbf{v}^{(L)}_i$ represents a $h \times w$ sub-patch. Hence, there is $N^{(L)} = \frac{H}{h} \cdot \frac{W}{w}$ local nodes. To define the set of neighbors for each local node \(\mathbf{v}^{(L)}_i\), the Euclidean distances between each local node \(\mathbf{v}^{(L)}_i\) and all remaining nodes are calculated. Then, the top \( K \) nodes most similar to \(\mathbf{v}^{(L)}_i\) are considered its neighbors \(\mathcal{N}(\mathbf{v}^{(L)}_i)\). After the set of neighbors for each local node is specified, the local edge set \(\mathcal{E}^{(L)}\) is constructed, where an edge $e_{i,j}$ exists if and only if $\mathbf{v}^{(L)}_j \in \mathcal{N}(\mathbf{v}^{(L)}_i)$. Then, the local-node network \( f^{(L)}_N \) learns to aggregate these neighbors to the local node \(\mathbf{v}^{(L)}_i\) as follow:

\begin{equation}
\mathbf{v}^{(L)'}_i = \sigma \Big( f^{(L)}_N\big(\mathbf{v}^{(L)}_i, \mathcal{E}^{(L)}\big) \Big),
\end{equation}

\noindent where $\sigma$ denotes nonlinear activation. To be specific, Max-relative graph convolution \citep{maxrelative} is adoped, followed by a projection layer to design $f^{(L)}_N$:

\begin{equation}
\begin{array}{ll}
f^{(L)}_N(\cdot) = F\Big(\big\lbrack \mathbf{v}^{(L)}_i; \max \big(\{\mathbf{v}^{(L)}_i - \mathbf{v}^{(L)}_j \vert e_{i,j} \in \mathcal{E}^{(L)}\}\big) \big\rbrack\Big),
\end{array}
\end{equation}

\noindent where $F$ is a linear projection layer, and $[\cdot ; \cdot]$ denotes the concatenation operation. In this manner, differences between a local node $\mathbf{v}^{(L)}_i$ and its neighbors are calculated and pooled by taking the maximum values per channel dimension. Then, it is concatenated with $\mathbf{v}^{(L)}_i$ and processed by a projection layer $F$ as an information-exchanging operation. To enhance feature diversity, two linear projections are applied to each local node: one before and one after fitting to \(f^{(L)}_N\). Additionally, we implement a skip connection to prevent gradient vanishing. To strengthen feature transformation, a feed-forward network is placed between local-graph layers. Finally, the set of local nodes $\mathcal{V}^{(L)}$ undergoes max pooling, selecting the highest value in each channel to form the global node $\mathbf{v}^{(G)}$.

\subsubsection{{Global-graph network.}} During training, a mini-batch of image samples is used to construct a global graph $\mathcal{G}^{(G)}$. The number of samples in a mini-batch is the number of global nodes $N^{(G)}$. Herein, we design global-graph network $GoG$ to process $\mathcal{G}^{(G)}$. Following \citep{graph}, we design two networks: the edge network $f^{(G)}_E$ and the node network $f^{(G)}_N$. The edge network $f^{(G)}_E$ is responsible for producing similarity scores between global nodes, while the node network $f^{(G)}_N$ aggregates all global nodes to the considered global node $\mathbf{v}^{(G)}_i$ based on computed similarities. First, the similarity score between $\mathbf{v}^{(G)}_i$ and $\mathbf{v}^{(G)}_j$, is computed:

\begin{equation}
\hat{a}^{(G)}_{i,j} = Sigmoid\Big(f^{(G)}_{E}\big(\big\Vert \mathbf{v}^{(G)}_i - \mathbf{v}^{(G)}_j \big\Vert\big)\Big),
\label{eq:4}
\end{equation}

\noindent where $Sigmoid$ denotes the Sigmoid function, and $\hat{a}^{(G)}_{i,j}$ denotes an element of the global-graph unnormalized affinity matrix $\hat{\mathcal{A}}^{(G)} \in \mathbb{R}^{N^{(G)} \times N^{(G)}}$. Then, self-connections of global nodes are added to $\hat{\mathcal{A}}$, and it is normalized:

\begin{equation}
\mathcal{A}^{(G)} = D^{-\frac{1}{2}}(\hat{\mathcal{A}}^{(G)} + I)D^{-\frac{1}{2}},
\label{eq:5}
\end{equation}

\noindent where $D$ is the degree matrix of $\hat{\mathcal{A}}^{(G)} + I$, $I$ is the identity matrix, and $\mathcal{A}^{(G)}$ is the normalized affinity matrix. The feature aggregation of the global node $\mathbf{v}^{(G)}_i$ is processed as follows:

\begin{equation}
\mathbf{v}^{(G)'}_i = f^{(G)}_N \big (\big\lbrack \mathbf{v}^{(G)}_i ; \sum_{j \in N^{(G)}} a^{(G)}_{i,j} \cdot \mathbf{v}^{(G)}_j \big\rbrack\big),
\label{eq:6}
\end{equation}

\noindent where $\mathbf{v}^{(G)'}_i$ is the updated feature, which is aggregated from its neighboring node representations. By doing so, all global nodes in $\mathcal{V}^{(G)}$ are multiplied by their similarity scores with the considered global node $\mathbf{v}^{(G)}_i$, followed by computing a weighted sum. This result is then concatenated with the global node $\mathbf{v}^{(G)}_i$ itself and passed through the node network $f^{(G)}_N$ as an update function.

\subsection{{Training scheme}}

\subsubsection{{Supervised learning.}} To perform supervised learning on labeled samples, we use the cross-entropy loss function to optimize $LoG$ and $GoG$:

\begin{equation}
\mathcal{L}_{node} = -\frac{1}{N^{(G)}}\sum^{N^{(G)}}_{i=1} y_i \log {p}(x_i),
\label{eq:10}
\end{equation}

\noindent where $p(x_i) = \softmax(GoG(LoG(x_i)))$.

\subsubsection{{Global edge supervision.}} Given the set of similarity scores $\{a^{(G)}_{i,j}\}^{N^{(G)}}_{j=1}$ between $\mathbf{v}^{(G)}_i$ and its neighbor nodes, produced by the global-edge network $f^{(G)}_E$, to guide the edge construction process, the binary cross-entropy loss is used as follows:

\begin{equation}
\mathcal{L}_{edge} = - e^{(G)}_{i,j} \log a^{(G)}_{i,j} - (1 - e^{(G)}_{i,j}) \log (1 - a^{(G)}_{i,j}),
\label{eq:12}
\end{equation}

\noindent where $e^{(G)}_{i,j}$ is the global ground-truth edge, $e^{(G)}_{i,j} = 1$ if and only if $\mathbf{v}^{(G)}_i$ and $\mathbf{v}^{(G)}_j$ belong to the same category; otherwise, $e^{(G)}_{i,j} = 0$.

\subsubsection{{Final objective function.}} Finally, the loss functions from Eq. \ref{eq:10} and Eq. \ref{eq:12} are merged into a unified objective function for training the whole network, which is formed as:

\begin{equation}
\mathcal{L}_{HiGDA} = \mathcal{L}_{node} + \mathcal{L}_{edge}.
\label{eq:higda}
\end{equation}

\subsection{{Graph Active Learning Strategy}} 
Due to the dominance of source samples, data bias may occur, leading to the suboptimal problem of the HiGDA on the target domain. Hence, we incorporate the concept of active learning to develop the graph active learning algorithm (GAL). In GAL, the pretraining-finetuning manner is repeatedly conducted in each training episode to select the most informative samples from the unlabeled target data for labeling aiming to optimize the model performance, which is processed as follows:

\noindent\textbf{Step 1.} In each episode, the pre-trained model on labeled samples is used to create pseudo labels from the unlabeled set $D_{u}$ as follows:
\begin{equation}
    \hat{y}^{i}_{ps}=\argmax p(x^{u}_{i})\; \; \text{if only if} \; \; \maxa (p(x^{u}_{i})) \geq \tau,
\end{equation}
\noindent with $p(x^{u}_{i})= \softmax(GoG(LoG(x^{u}_{i})))$, and $\tau$ is the predefined threshold. The pseudo label set is denoted as $D_{ps}=\{x^{u}_{i}, \hat{y}^{i}_{ps}\}^{N_{ps}}_{i=1}$, where $N_{ps}$ is the number of pseudo labels.

\noindent\textbf{Step 2.} Then, we update the labeled set $D_{l}$ by combining it with the pseudo label set $D_{ps}$ every episode as follows:
\begin{equation}
    \hat{D}_{l}^{q+1} = D_{l}\cup D_{ps}^{q}, 
\end{equation} 
\noindent where $D_{ps}^{q}$ is the pseudo label set at the training episode $q$. The HiGDA model is then fine-tuned with the graph active learning strategy to get the optimal model performance on the updated version $\hat{D}_{l}$ with the supervised losses in Eq. (8) and Eq. (9). The pairwise connection rule to determine edges among samples in $\hat{D}_{l}$ for feature aggregation is defined as follows:
\begin{equation} 
e^{(G)}_{i,j} = \begin{cases}
1,&{\text{if}}\ y^{l}_{i}=y^{l}_{j},  \\ 
1,&{\text{if}}\ {y}^{l}_{i}=\hat{y}^{ps}_{j}, \\
1,&{\text{if}}\ \hat{y}^{ps}_{i}=\hat{y}^{ps}_{j}, \\
{0,}&{\text{otherwise.}} 
\end{cases}
\end{equation}

In this manner, the performance of HiGDA can be adjusted according to the quality and quantity of generated pseudo labels.

\vspace{-0.2cm}

\section{Integrating HiGDA with SSDA Techniques} 

In this section, we deploy the proposed framework HiGDA with the previous SSDA method.

\noindent\textbf{Minimax Entropy (MME).} MME \citep{MME} is the most popular SSDA method using adversarial learning with a minimax entropy strategy expressed as follows:

\begin{equation}
\begin{aligned}
{\theta}^{*}_{LoG} & = \argmin\limits_{\theta_{LoG}}\mathcal{L}_{HiGDA} + \lambda\mathcal{L}_{ENT}, \\
{\theta}^{*}_{GoG} & = \argmin\limits_{\theta_{GoG}}\mathcal{L}_{HiGDA} - \lambda\mathcal{L}_{ENT},
\end{aligned}
\end{equation}
\noindent where $\mathcal{L}_{HiGDA}$ is the supervised learning loss in Eq. (\ref{eq:higda}) calculated using the labeled samples, $\mathcal{L}_{ENT}$ is the entropy loss estimated by using the unlabeled target samples. $\lambda = 0.1$ is the trade-off hyperparameter.

\noindent\textbf{Adversarial Adaptive Clustering (AAC).} Similar to MME, AAC \citep{AAC} also uses adversarial learning between a feature extractor and a classifier by proposing the new adversarial adaptive clustering loss to estimate the pairwise feature similarity among unlabeled target data within a mini-batch, as follows: 


\begin{equation}
\begin{aligned}
{\theta}^{*}_{LoG} & = \argmin\limits_{\theta_{LoG}}\mathcal{L}_{HiGDA} + \beta\mathcal{L}_{AAC}, \\
{\theta}^{*}_{GoG} & = \argmin\limits_{\theta_{GoG}}\mathcal{L}_{HiGDA} - \beta\mathcal{L}_{AAC},
\end{aligned}
\end{equation}
\noindent where $\mathcal{L}_{AAC}$ is the adversarial adaptive clustering loss with $\beta$ to be 1.0. 

\begin{table*}[h]
\centering
\resizebox{0.8\textwidth}{!}{%
\begin{tabular}{l|l|cccccccccccc|c} \toprule
\multicolumn{1}{c|}{Setting} & Method & {R$\rightarrow$C} & {R$\rightarrow$P} & {R$\rightarrow$A} & {P$\rightarrow$R} & {P$\rightarrow$C} & {P$\rightarrow$A} & {A$\rightarrow$P} & {A$\rightarrow$C} & {A$\rightarrow$R} & {C$\rightarrow$R} & {C$\rightarrow$A} & {C$\rightarrow$P} & Avg. (\%) \\ \midrule
\multicolumn{1}{c|}{ } & \multicolumn{14}{c}{\cellcolor[HTML]{DCDCDC}ResNet-34}  \\ \cmidrule{2-15} 
\multirow{18}{*}{\rotatebox[origin=c]{90}{1-shot}} & S+T   (Baseline) & 52.1 & 78.6 & 66.2 & 74.4 & 48.3 & 57.2 & 69.8 & 50.9 & 73.8 & 70.0 & 56.3 & 68.1 & 63.8 \\
 & MME (ICCV'19) & 61.9 & 82.8 & 71.2 & 79.2 & 57.4 & 64.7 & 75.5 & 59.6 & 77.8 & 74.8 & 65.7 & 74.5 & 70.4 \\
 
 & CDAC (CVPR'21) & 61.9 & 83.1 & 72.7 & 80.0 & 59.3 & 64.6 & 75.9 & 61.2 & 78.5 & 75.3 & 64.5 & 75.1 & 71.0 \\
 
 & DECOTA (ICCV'21) & 56.0 & 79.4 & 71.3 & 76.9 & 48.8 & 60.0 & 68.5 & 42.1 & 72.6 & 70.7 & 60.3 & 70.4 & 64.8 \\

 & MCL (IJCAI'22) & 67.0 & 85.5 & 73.8 & 81.3 & 61.1 & 68.0 & 79.5 & 64.4 & 81.2 & 78.4 & 68.5 & 79.3 & 74.0 \\
 
 & MME + SLA (CVPR'23) & 64.1 & 83.8 & 72.9 & 80.0 & 59.9 & 66.7 & 76.3 & 62.1 & 78.6 & 75.1 & 67.5 & 77.1 & 72.0 \\

 & SPA (NeurIPS'23) & 65.2 & 84.1 & 71.4 & 80.7 & 59.1 & 65.7 & 76.7 & 62.3 & 79.0 & 76.4 & 66.6 & 77.3 & 72.0 \\

 & FixMME + EFTL (AAAI'24) & 66.6 & 87.2 & 74.3 & 82.6 & 63.3 & 68.7 & 80.5 & 65.7 & 80.8 & 77.5 & 65.6 & 79.6 & 74.4 \\ 

 \cmidrule{2-15}
\multicolumn{1}{c|}{ } & \multicolumn{14}{c}{\cellcolor[HTML]{FFFACD}HiGDA-T} \\ \cmidrule{2-15}

 & S+T (Baseline) & 57.9 & 86.9 & 77.3 & 81.9 & 52.4 & 63.8 & 76.4 & 54.7 & 80.0 & 76.3 & 62.6 & 71.2 & 70.1 \\
 
 & Baseline+AAC & 67.7 & 87.3 & 84.1 & 87.6 & 62.9 & 75.4 & 78.6 & 61.8 & 86.7 & 83.5 & 71.3 & 79.1 & 77.2 \\ 

 & Baseline+MME & 70.4 & 92.0 & 88.1 & 90.5 & 65.0 & 78.9 & 84.2 & 66.1 & 89.5 & 86.4 & 76.4 & 83.2 & 80.9 \\
 \cmidrule{2-15}

 \multicolumn{1}{c|}{ } & \multicolumn{14}{c}{\cellcolor[HTML]{FFE4E1}HiGDA-T+GAL} \\ \cmidrule{2-15}

 & S+T (Baseline) & 66.7 & 87.3 & 79.9 & 85.0 & 59.8 & 71.1 & 78.8 & 62.1 & 79.1 & 79.6 & 72.2 & 73.8 & 74.6 \\
 
 & Baseline+AAC & 81.3 & 91.1 & 87.7 & 92.6 & 77.9 & 83.2 & 85.1 & 76.5 & 90.0 & 88.6 & 82.6 & 85.4 & 85.2 \\  

 & Baseline+MME & \textbf{82.5} & \textbf{93.9} & \textbf{90.4} & \textbf{92.7} & \textbf{79.3} & \textbf{84.3} & \textbf{86.8} & \textbf{80.2} & \textbf{94.1} & \textbf{91.6} & \textbf{87.3} & \textbf{86.5} & \textbf{87.5} \\
 \midrule

\multicolumn{1}{c|}{ } & \multicolumn{14}{c}{\cellcolor[HTML]{DCDCDC}ResNet-34} \\ \cmidrule{2-15}

\multirow{18}{*}{\rotatebox[origin=c]{90}{3-shot}} & S+T (Baseline) & 55.7 & 80.8 & 67.8 & 73.1 & 53.8 & 63.5 & 73.1 & 54.0 & 74.2 & 68.3 & 57.6 & 72.3 & 66.2 \\

 
 & MME (ICCV'19) & 64.6 & 85.5 & 71.3 & 80.1 & 64.6 & 65.5 & 79.0 & 63.6 & 79.7 & 76.6 & 67.2 & 79.3 & 73.1 \\
 
 & CDAC (CVPR'21) & 67.8 & 85.6 & 72.2 & 81.9 & 67.0 & 67.5 & 80.3 & 65.9 & 80.6 & 80.2 & 67.4 & 81.4 & 74.2 \\
 
 & DECOTA (ICCV'21) & 70.4 & 87.7 & 74.0 & 82.1 & 68.0 & 69.9 & 81.8 & 64.0 & 80.5 & 79.0 & 68.0 & 83.2 & 75.7 \\

 & MCL (IJCAI'22) & 70.1 & 88.1 & 75.3 & 83.0 & 68.0 & 69.9 & 83.9 & 67.5 & 82.4 & 81.6 & 71.4 & 84.3 & 77.1 \\

 & SPA (NeurIPS'23) & 67.2 & 87.0 & 73.9 & 82.0 & 65.2 & 69.5 & 81.0 & 63.1 & 80.2 & 77.5 & 68.5 & 81.7 & 74.7 \\
 
 & MME + SLA (CVPR'23) & 68.4 & 87.4 & 74.7 & 81.9 & 67.4 & 69.7 & 81.1 & 65.9 & 80.5 & 79.4 & 69.2 & 81.9 & 75.6 \\

 & FixMME + EFTL (AAAI'24) & 72.8 & 89.3 & 77.5 & 85.4 & 70.9 & 72.6 & 84.8 & 70.3 & 83.8 & 81.5 & 70.6 & 84.6 & 78.7 \\ 

  
 \cmidrule{2-15}
 
\multicolumn{1}{c|}{ } & \multicolumn{14}{c}{\cellcolor[HTML]{FFFACD}HiGDA-T} \\ \cmidrule{2-15}
 & S+T (Baseline) & 64.7 & 90.3 & 78.2 & 84.8 & 60.0 & 69.8 & 82.6 & 58.4 & 83.1 & 71.9 & 70.0 & 80.4 & 74.5 \\
 

 & Baseline+AAC & 73.1 & 91.0 & 84.1 & 89.7 & 71.1 & 79.8 & 85.5 & 66.9 & 89.6 & 85.8 & 77.2 & 86.0 & 81.7 \\ 

 & Baseline+MME & 76.8 & 94.1 & 88.3 & 92.6 & 72.7 & 82.4 & 89.8 & 73.9 & 92.5 & 90.1 & 80.8 & 88.7 & 85.2 \\
 \cmidrule{2-15}

 \multicolumn{1}{c|}{ } & \multicolumn{14}{c}{\cellcolor[HTML]{FFE4E1}HiGDA-T+GAL} \\ \cmidrule{2-15}

 & S+T (Baseline) & 71.0 & 92.1 & 81.5 & 87.1 & 66.4 & 74.8 & 86.2 & 66.2 & 86.5 & 84.3 & 74.2 & 87.8 & 79.8 \\
 
 & Baseline+AAC & 84.6 & 94.9 & 88.8 & 94.1 & 80.7 & 85.3 & 91.2 & 81.1 & 94.9 & 92.7 & 86.8 & 91.2 & 88.9 \\ 

 & Baseline+MME & \textbf{84.8} & \textbf{95.1} & \textbf{91.8} & \textbf{94.3} & \textbf{84.2} & \textbf{88.4} & \textbf{92.3} & \textbf{81.6} & \textbf{95.8} & \textbf{93.9} & \textbf{87.6} & \textbf{92.5} & \textbf{90.2} \\
 \bottomrule

\end{tabular}
} \vspace{-0.2cm}
\caption{Classification accuracy (\%) on \textit{Office-Home} for 1-shot and 3-shot settings.}
\label{tab:officehome} \vspace{-0.4cm}
\end{table*}

\vspace{-0.3cm}
\section{Experiments}
We evaluate our framework on various challenging domain adaptation datasets, including \textit{Office-Home} \citep{Office-Home}, \textit{DomainNet} \citep{DomainNet} and \textit{VisDA2017} \citep{VisDA2017} in Tabs. \ref{tab:officehome}, \ref{tab:domainnet} and \ref{tab:visda}, respectively. Consistent with previous SSDA methods, we perform 12 DA tasks on \textit{Office-Home} and 7 DA tasks on \textit{DomainNet} under 1-shot and 3-shot settings, while considering only one domain adaptation scenario for \textit{VisDA2017}. 
\vspace{-0.2cm}

\subsection{Implemental Details} 

For fairness to the CNN-based network in terms of the number of parameters, we select the tiny version of Pyramid ViG \citep{VisionGNN} for $LoG$, where the number of local nodes $N^{(L)}$ obtained from the last block is $49$. The number of neighbors for each local node is set to $9$. We are inspired by \citep{graph} to build $GoG$ consisting of the node and edge networks. The number of global nodes $N^{(G)}$ is set to $32$. For training, we use the SGD optimizer, the learning rate and weight decay are set to $1\times 10^{-3}$ and $5\times 10^{-5}$, respectively. The threshold $\tau$ in Eq. (9) is set to $0.95$ and the episode in Eq. (10) is set to $q=50$ consisting of $1,000$ training steps. A single NVIDIA GeForce RTX 4090 GPU is used for all experiments.
\vspace{-0.2cm}
\subsection{Comparison Results}

As shown in Tabs.~\ref{tab:officehome}, \ref{tab:domainnet}, and \ref{tab:visda}, HiGDA-T baseline significantly outperforms the CNN-based baseline model with the same S+T setting. Surprisingly, the baseline can also surpass the existing CNN-based SSDA methods. Specifically, on \textit{Office-Home} in Tab. \ref{tab:officehome}, HiGDA-T exceeds DECOTA by 5.3\% in the 1-shot setting, and overcomes MME \citep{MME} and CDAC \citep{AAC} by 1.4\% and 0.3\% in the 3-shot setting, respectively. 
Furthermore, as in Tab. \ref{tab:domainnet} on \textit{DomainNet}, when HiGDA-T is enhanced with the proposed GAL, referred to as HiGDA-T+GAL, it shows marginal improvements over the second-best method, EFTL \citep{EFTL}, with increasing of up to 5.1\% and 4.8\% under the 1-shot and 3-shot settings, respectively. The proposed method achieves the highest performance with implementation of HiGDA-T+GAL+MME, surpassing EFTL \citep{EFTL} by as much as 15.2\% and 15.8\% with 1- and 3-shot settings, respectively. Similarly, our method reaches 97.8\% on \textit{VisDA2017} with  HiGDA-T+GAL+MME under 3-shot setting, as in Tab. \ref{tab:visda}.

\subsection{Analyses} To further validate HiGDA, we investigate three key aspects as \textit{ablation studies}: (1) the ability of HiGDA to integrate with existing SSDA methods; (2) the effectiveness of the proposed GAL on cross-domain alignment; and (3) the effectiveness of $LoG$ compared to other CNN-based and ViT-based backbones, as well as the performance of $GoG$ compared to MLP.

\begin{table*}
\parbox{.74\linewidth}{
\centering
\resizebox{0.7\textwidth}{!}{
\begin{tabular}{l|cccccccccccccc|cc} \toprule
\multirow{2}{*}{Method} & 
\multicolumn{2}{c}{rel$\rightarrow$clp} & 
\multicolumn{2}{c}{rel$\rightarrow$pnt} & 
\multicolumn{2}{c}{pnt$\rightarrow$clp} & 
\multicolumn{2}{c}{clp$\rightarrow$skt} & 
\multicolumn{2}{c}{skt$\rightarrow$pnt} & 
\multicolumn{2}{c}{rel$\rightarrow$skt} & 
\multicolumn{2}{c|}{pnt$\rightarrow$rel} & 
\multicolumn{2}{c}{Avg. (\%)} \\
& \multicolumn{1}{l}{1\textsubscript{shot}} &
\multicolumn{1}{l}{3\textsubscript{shot}} &
\multicolumn{1}{l}{1\textsubscript{shot}} &
\multicolumn{1}{l}{3\textsubscript{shot}} &
\multicolumn{1}{l}{1\textsubscript{shot}} &
\multicolumn{1}{l}{3\textsubscript{shot}} &
\multicolumn{1}{l}{1\textsubscript{shot}} &
\multicolumn{1}{l}{3\textsubscript{shot}} &
\multicolumn{1}{l}{1\textsubscript{shot}} &
\multicolumn{1}{l}{3\textsubscript{shot}} &
\multicolumn{1}{l}{1\textsubscript{shot}} &
\multicolumn{1}{l}{3\textsubscript{shot}} &
\multicolumn{1}{l}{1\textsubscript{shot}} &
\multicolumn{1}{l|}{3\textsubscript{shot}} &
\multicolumn{1}{l}{1\textsubscript{shot}} &
\multicolumn{1}{l}{3\textsubscript{shot}}  \\ \midrule
\multicolumn{17}{c}{\cellcolor[HTML]{DCDCDC}ResNet-34}  \\ \midrule
S+T (Baseline) & 55.6 & 60.0 & 60.6 & 62.2 & 56.8 & 59.4 & 50.8 & 55.0 & 56.0 & 59.5 & 46.3 & 50.1 & 71.8 & 73.9 & 56.8 & 60.0 \\

MME (ICCV'19) & 70.0 & 72.2 & 67.7 & 69.7 & 69.0 & 71.7 & 56.3 & 61.8 & 64.8 & 66.8 & 61.0 & 61.9 & 76.1 & 78.5 & 66.4 & 68.9  \\

DECOTA (ICCV'21) & 79.1 & 80.4 & 74.9 & 75.2 & 76.9 & 78.7 & 65.1 & 68.6 & 72.0 & 72.7 & 69.7 & 71.9 & 79.6 & 81.5 & 73.9 & 75.6 \\

CDAC (CVPR'21) & 77.4 & 79.6 & 74.2 & 75.1 & 75.5 & 79.3 & 67.6 & 69.9 & 71.0 & 73.4 & 69.2 & 72.5 & 80.4 & 81.9 & 73.6 & 76.0 \\

MCL (IJCAI'22) & 77.4 & 79.4 & 74.6 & 76.3 & 75.5 & 78.8 & 66.4 & 70.9 & 74.0 & 74.7 & 70.7 & 72.3 & 82.0 & 83.3 & 74.4 & 76.5 \\

SPA (NeurIPS'23) & 75.3 & 76.0 & 71.8 & 72.2 & 74.8 & 76.5 & 65.9 & 67.0 & 69.7 & 71.1 & 65.8 & 67.2 & 81.1 & 82.3 & 72.1 & 73.2 \\

CDAC + SLA (CVPR’23) & 79.8 & 81.6 & 75.6 & 76.0 & 77.4 & 90.3 & 68.1 & 71.3 & 71.7 & 73.5 & 71.7 & 73.5 & 80.4 & 82.5 & 75.0 & 76.9  \\

FixMME + EFTL (AAAI’24) & 79.6 & 81.2 & 74.9 & 77.1 & 78.2 & 81.8 & 69.3 & 72.8 & 71.8   & 74.4   & 69.9 & 71.5 & 83.1 & 84.4 & 75.3 & 77.6 \\ \midrule

\multicolumn{17}{c}{\cellcolor[HTML]{FFFACD}HiGDA-T}   \\ \midrule
S+T (Baseline) & 68.8  & 75.3  & 74.3  & 78.3 & 69.4 & 75.6 & 61.0 & {71.1} & 68.6 & 73.9 & 60.2 & 65.6 & 82.2 & 87.0 & 69.2 & {75.3} \\

Baseline+AAC & 80.2 & 84.7 & 86.4 & 87.6 & 80.5 & 84.8 & 76.2 & 79.8 & 84.2 & 86.2 & 75.1 & 77.9 & 90.1 & 92.5 & 81.8 & 84.8 \\ 

Baseline+MME & 81.5 & 85.7 & 88.0 & 89.7 & 84.7 & 88.4 & 77.9 & 80.4 & 85.4 & 87.8 & 75.6 & 79.1 & 90.9 & 93.0 & 83.5 & 86.4 \\ 
\midrule

\multicolumn{17}{c}{\cellcolor[HTML]{FFE4E1}HiGDA-T + GAL}  \\ \midrule
S+T (Baseline) & 78.5 & 85.1 & 80.9 & 84.6 & 73.1 & 83.7 & 64.5 & 76.0 & 77.6 & 80.4 & 71.9 & 78.3 & 87.8 & 88.6 & 76.3 & 82.4 \\

Baseline+AAC & 90.9 & 92.7 & 92.4 & 94.2 & 88.4 & 93.5 & 85.2 & 89.3 & 90.6 & 93.8 & 89.7 & 90.9 & 90.2 & 94.3 & 89.6 & 92.7 \\ 

Baseline+MME & \textbf{92.1} & \textbf{93.9} & \textbf{93.8} & \textbf{95.0} & \textbf{89.6} & \textbf{95.4} & \textbf{85.7} & \textbf{89.7} & \textbf{91.2} & \textbf{94.0} & \textbf{90.1} & \textbf{91.5} & \textbf{91.0} & \textbf{94.6} & \textbf{90.5} & \textbf{93.4} \\

\bottomrule

\end{tabular}}
\caption{Classification accuracy (\%) on \textit{DomainNet} for 1-shot and 3-shot settings.}
\label{tab:domainnet}
}
\hfill
\parbox{.23\linewidth}{
\centering
\resizebox{0.21\textwidth}{!}{
\begin{tabular}{l|cc} \toprule
Method & \multicolumn{1}{l}{1\textsubscript{shot}} &
\multicolumn{1}{l}{3\textsubscript{shot}} \\ \midrule
\multicolumn{3}{c}{\cellcolor[HTML]{DCDCDC}ResNet-34}  \\ \midrule
S+T (Baseline) & 60.2 & 64.6 \\
MME (ICCV'19) & 68.7 & 70.9 \\
APE (ECCV'20) & 78.9 & 81 \\
CDAC (CVPR'21) & 69.9 & 80.6 \\
DECOTA (ICCV'21) & 64.9 & 80.7 \\
ECACL (ICCV'21)  & 81.1 & 83.3 \\
MCL (IJCAI'22) & 86.3 & 87.3 \\
ProML (IJCAI'23)  & 87.6 & 88.4 \\ \midrule

\multicolumn{3}{c}{\cellcolor[HTML]{FFE4E1}HiGDA-T + GAL}  \\ \midrule
S+T (Baseline) & 80.9 & 93.6 \\
Baseline+AAC & \textbf{91.4} & 96.3 \\
Baseline+MME & 88.9 & \textbf{97.8} \\ 
\bottomrule
\end{tabular}
}
\caption{Mean class-wise accuracy (\%) on \textit{VisDA2017} for 1-shot and 3-shot settings.}
\label{tab:visda}
}
\end{table*}

\begin{table*}[]
\centering
\resizebox{0.7\textwidth}{!}{%
\begin{tabular}{c|c|c|ccccccc|c} \toprule
Feature Extractor & Classifier & Parameter (M) & rel$\rightarrow$clp & rel$\rightarrow$pnt & pnt$\rightarrow$clp & clp$\rightarrow$skt & skt$\rightarrow$pnt & rel$\rightarrow$skt & pnt$\rightarrow$rel & Avg.   (\%) \\ \midrule
ResNet-34 &  & 21.6 & 60.6 & 63.0 & 62.8 & 54.9 & 62.0 & 51.5 & 76.5 & 61.6  \\

PVT-Tiny &  & 13.8 & 56.9 & 60.0 & 57.2  & 49.6 & 57.1 & 45.3 & 76.1 & 57.5 \\

$LoG$-Tiny & \multirow{-3}{*}{MLP} & 9.5 & 64.7 & 66.9 & 64.5 & 56.5 & 64.9 & 54.3 & 79.8 & 64.5 \\ \midrule

ResNet-34 & & 22.6 & 73.1 & 75.3 & 74.1 & 67.1 & 73.4 & 64.0 & 84.1 & 73.0 \\

PVT-Tiny & & 14.8 & 65.0 & 67.4 & 64.7 & 54.9 & 62.5 & 52.7 & 82.0 & 64.2 \\

$LoG$-Tiny  & \multirow{-3}{*}{$GoG$} & 10.5 & \textbf{75.3} & \textbf{78.3} & \textbf{75.6} & {\textbf{71.1}} & \textbf{73.9} & \textbf{65.6} & \textbf{87.0} & {\textbf{75.3}} \\ \bottomrule  
\end{tabular}
}
\caption{Ablation study on \textit{DomainNet} under the 3-shot setting to evaluate the effectiveness of different classifiers.}
\label{tab:log_gog} \vspace{-0.35cm}
\end{table*}

\subsubsection{Effectiveness of HiGDA on SSDA methods (1).} 
Results reported in Tabs. \ref{tab:officehome} and \ref{tab:domainnet}, highlighted under \colorbox[HTML]{FFFACD}{yellow} cells, demonstrate that HiGDA can be flexibly coupled with existing SSDA methods. Under the 3-shot setting, integrating HiGDA-T with AAC and MME improves performance by 7.2\% and 10.7\% compared to only the HiGDA-T baseline on \textit{Office-Home}, respectively. For \textit{DomainNet}, these improvements are 9.5\% and 11.1\%, respectively.

\subsubsection{Effectiveness of Graph Active Learning (2).} 
We report the impact of GAL (HiGDA-T+GAL)—highlighted under \colorbox[HTML]{FFE4E1}{pink} cells, by comparing it to the HiGDA baseline model—highlighted under \colorbox[HTML]{FFFACD}{yellow} cells. As listed in Tabs \ref{tab:officehome} and \ref{tab:domainnet}, the results show significant performance gaps when GAL is removed. Under the S+T setting with 3-shot, there is a performance drop of 5.3\% on \textit{Office-Home} and 7.1\% on \textit{DomainNet}. These results reveal that data bias still exists when only a few target labeled samples are available, and the proposed GAL can effectively address this issue by providing additional pseudo labels from unlabeled target samples.

\vspace{-0.15cm}
\subsubsection{Effectiveness of local and global graphs (3).} 

In the context of the \textit{feature-level}, $LoG$ significantly outperforms ResNet-34 \citep{ResNet-34} and PVT-Tiny \citep{PVT} across all evaluated head classifiers, as detailed in Table \ref{tab:log_gog}. Specifically, $LoG$ paired with the MLP head classifier demonstrates an increase in accuracy of 2.9\% and 7.0\% compared to ResNet-34 and PVT-Tiny, respectively, while utilizing fewer parameters. Similarly, when employing the $GoG$ as the classifier, $LoG$ also exceeds ResNet-34 and PVT-Tiny by 2.3\% and 11.1\%, respectively.
Regarding the \textit{category level}, $GoG$ consistently demonstrates enhanced performance compared to MLP, independent of the various backbones utilized. Notably, ResNet-34 equipped with $GoG$ yields the most significant improvement over the MLP, with an increase of 11.4\%. Finally, the highest overall performance of 75.3\% is attained by integrating $LoG$ with $GoG$ (HiGDA-T), thereby proving the effectiveness and robustness of HiGDA design. \vspace{-0.2cm}

\begin{figure}[!htt]
\resizebox{\columnwidth}{!}{%
\centering
    \begin{adjustbox}{width=0.44\textwidth,center}
    \begin{tabular}{cc|cc} \toprule
     \multicolumn{2}{c|}{\textbf{Dolphin}} & \multicolumn{2}{c}{\textbf{Pineapple}} \\ \midrule
     \multicolumn{1}{c}{\frame{\includegraphics[width=0.3\linewidth]{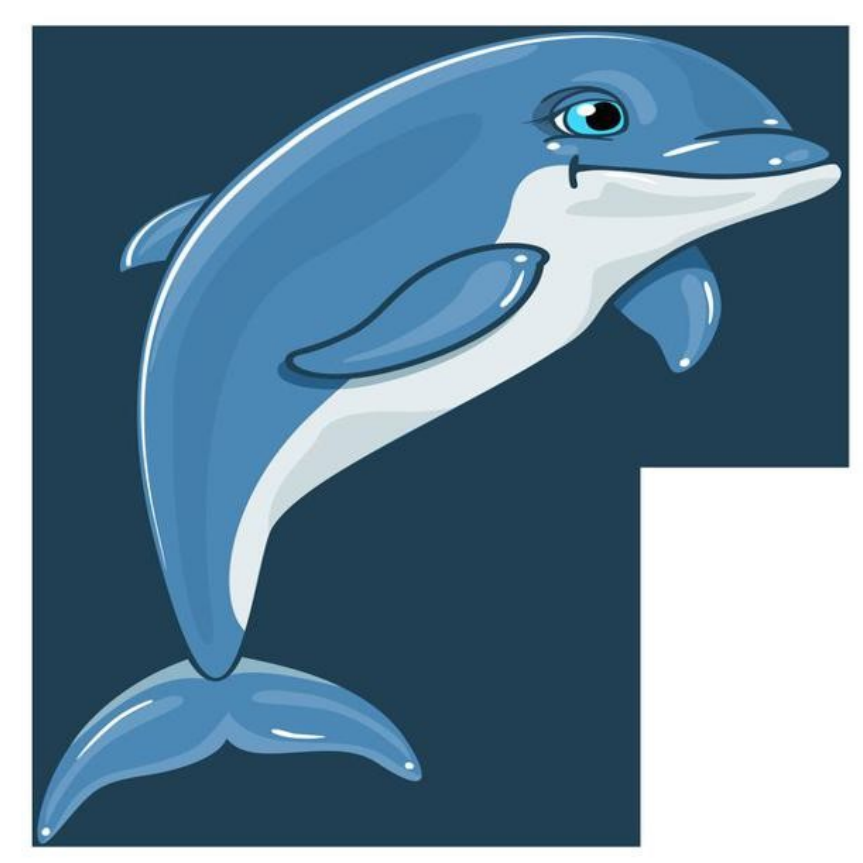}}} & \multicolumn{1}{c|}{\frame{\includegraphics[width=0.3\linewidth]{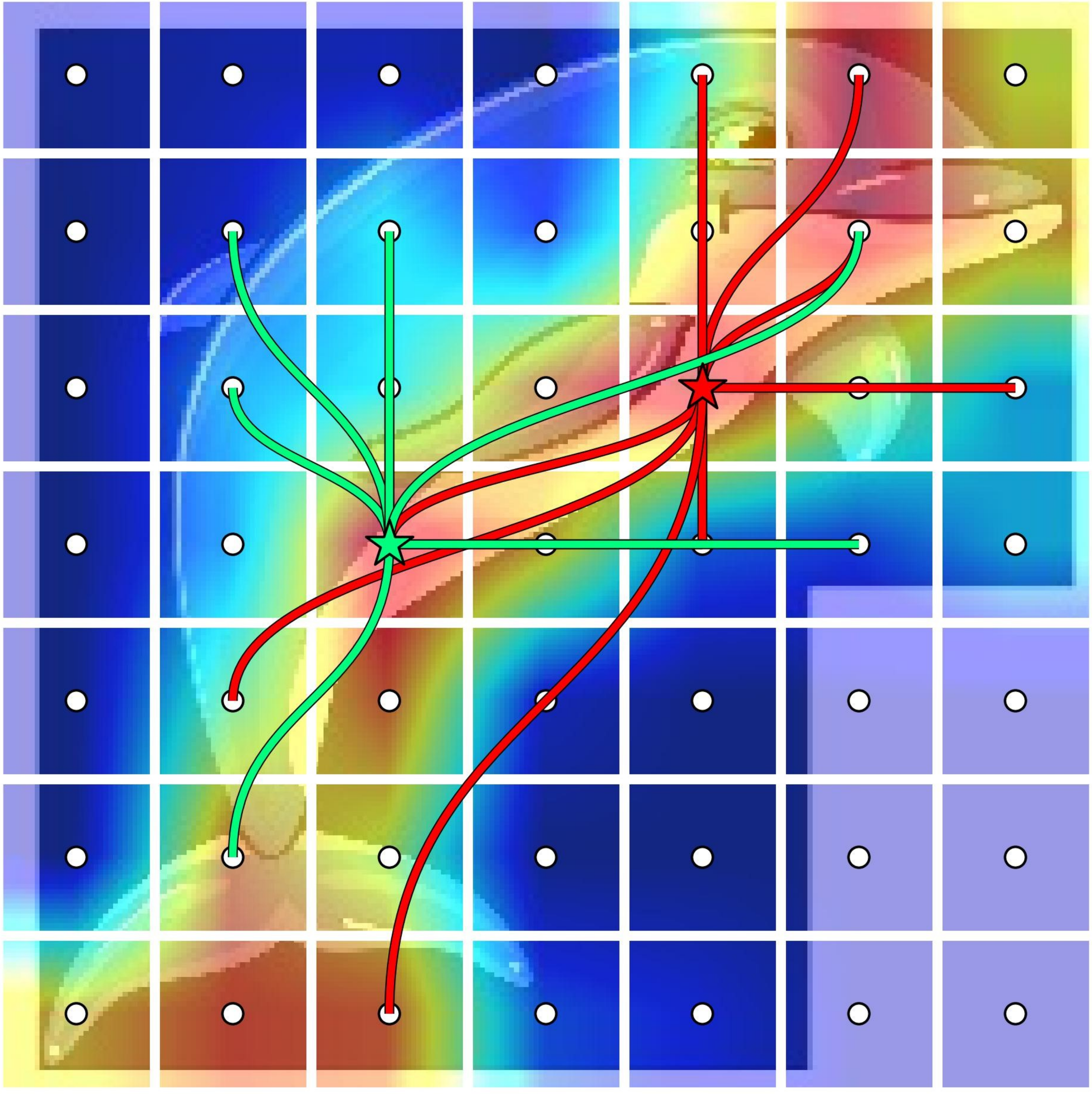}}} & \multicolumn{1}{c}{\frame{\includegraphics[width=0.3\linewidth]{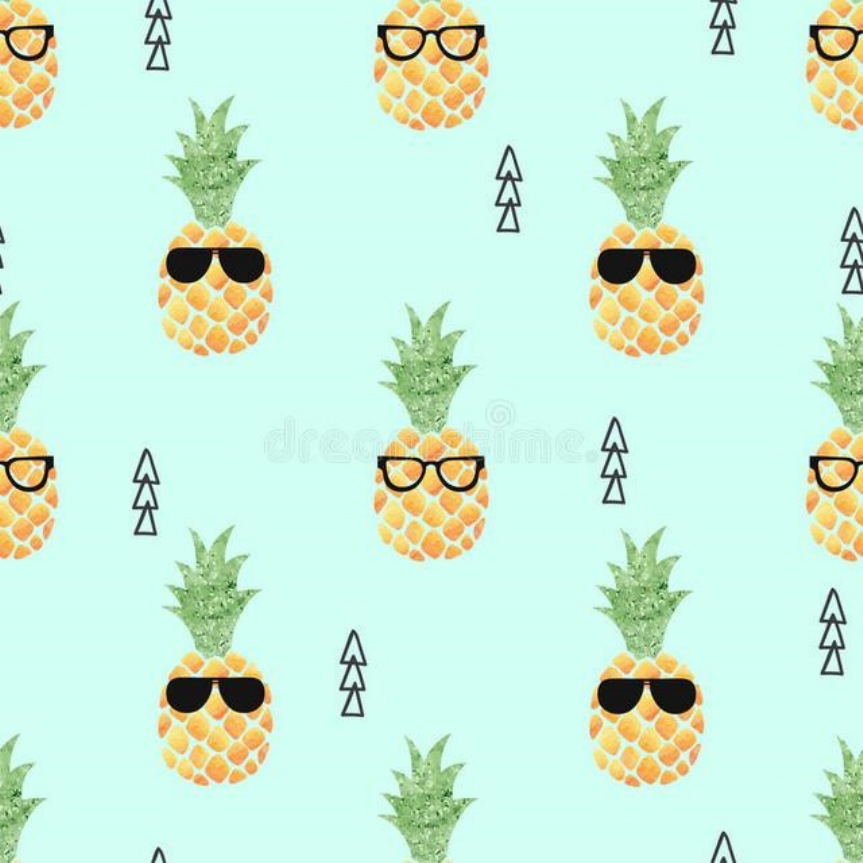}}} & \multicolumn{1}{c}{\frame{\includegraphics[width=0.3\linewidth]{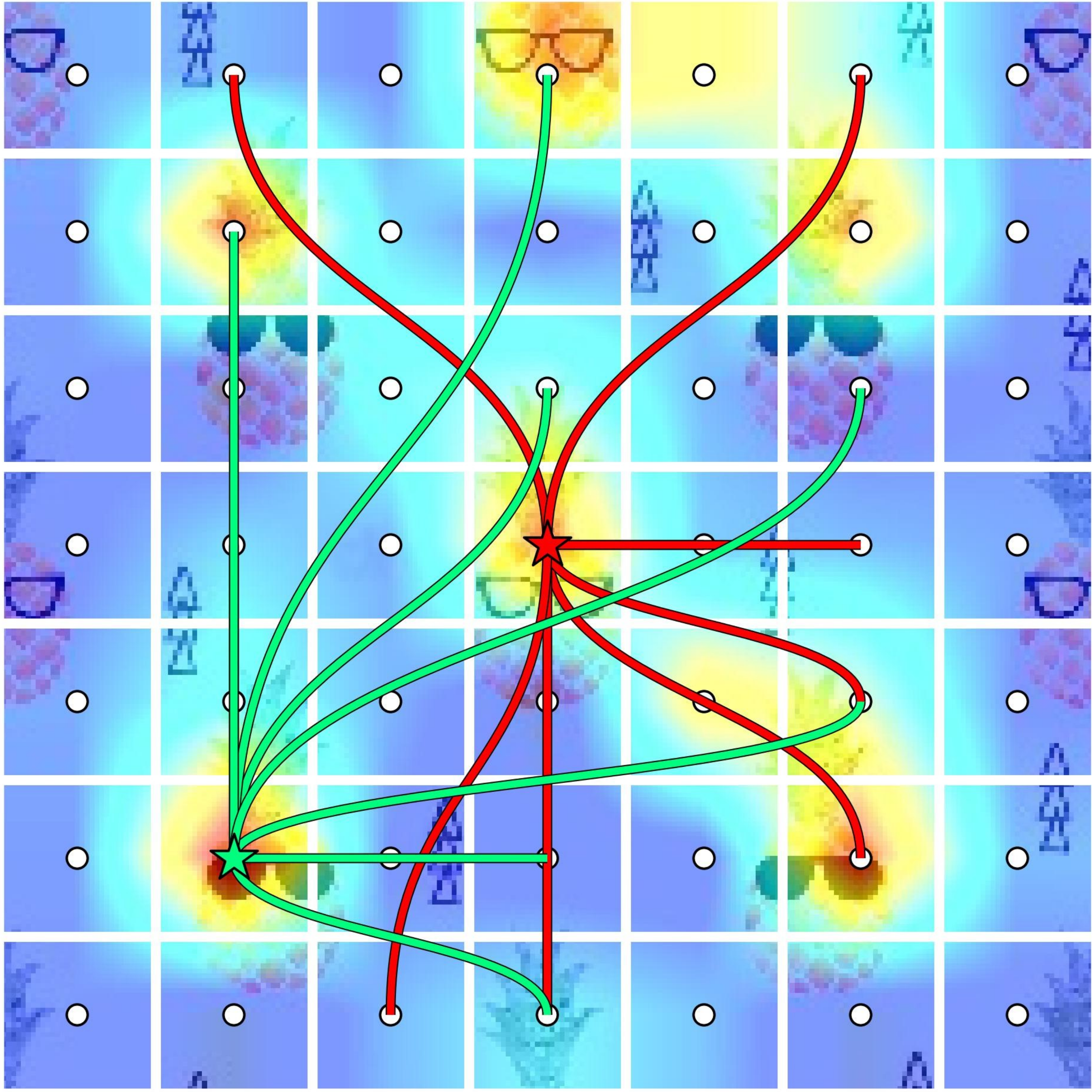}}} \\ \midrule
     \multicolumn{2}{c|}{(a) Single object} & \multicolumn{2}{c}{(b) Multiple objects} \\ \midrule
     \multicolumn{2}{c|}{\textbf{Horse}} & \multicolumn{2}{c}{\textbf{Strawberry}} \\ \midrule
     \multicolumn{1}{c}{\frame{\includegraphics[width=0.3\linewidth]{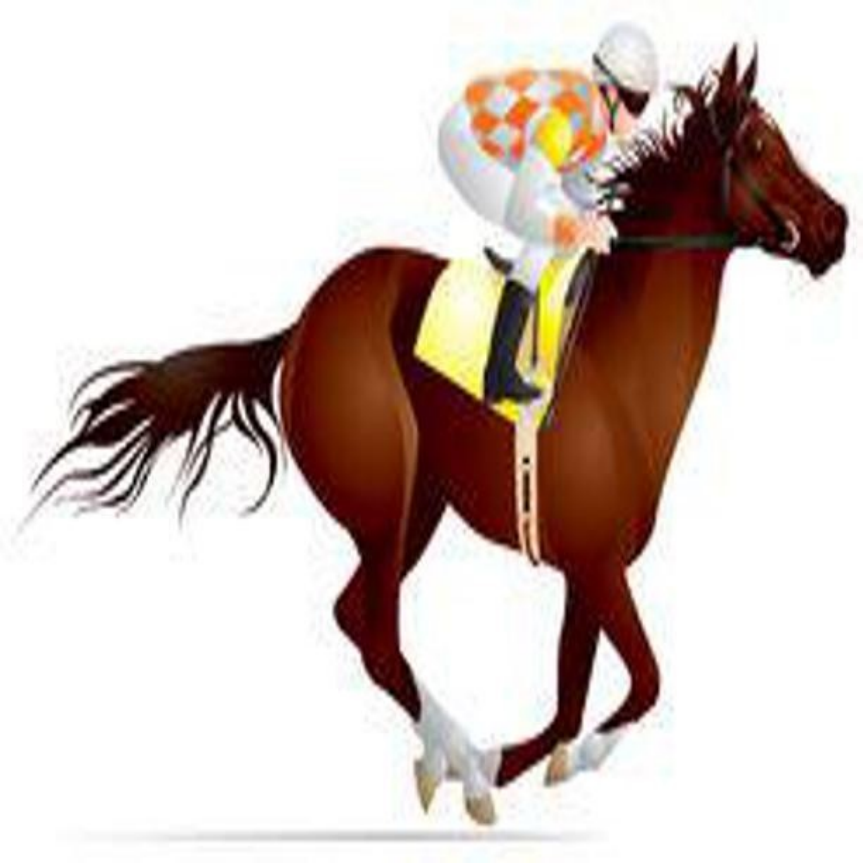}}} & \multicolumn{1}{c|}{\frame{\includegraphics[width=0.3\linewidth]{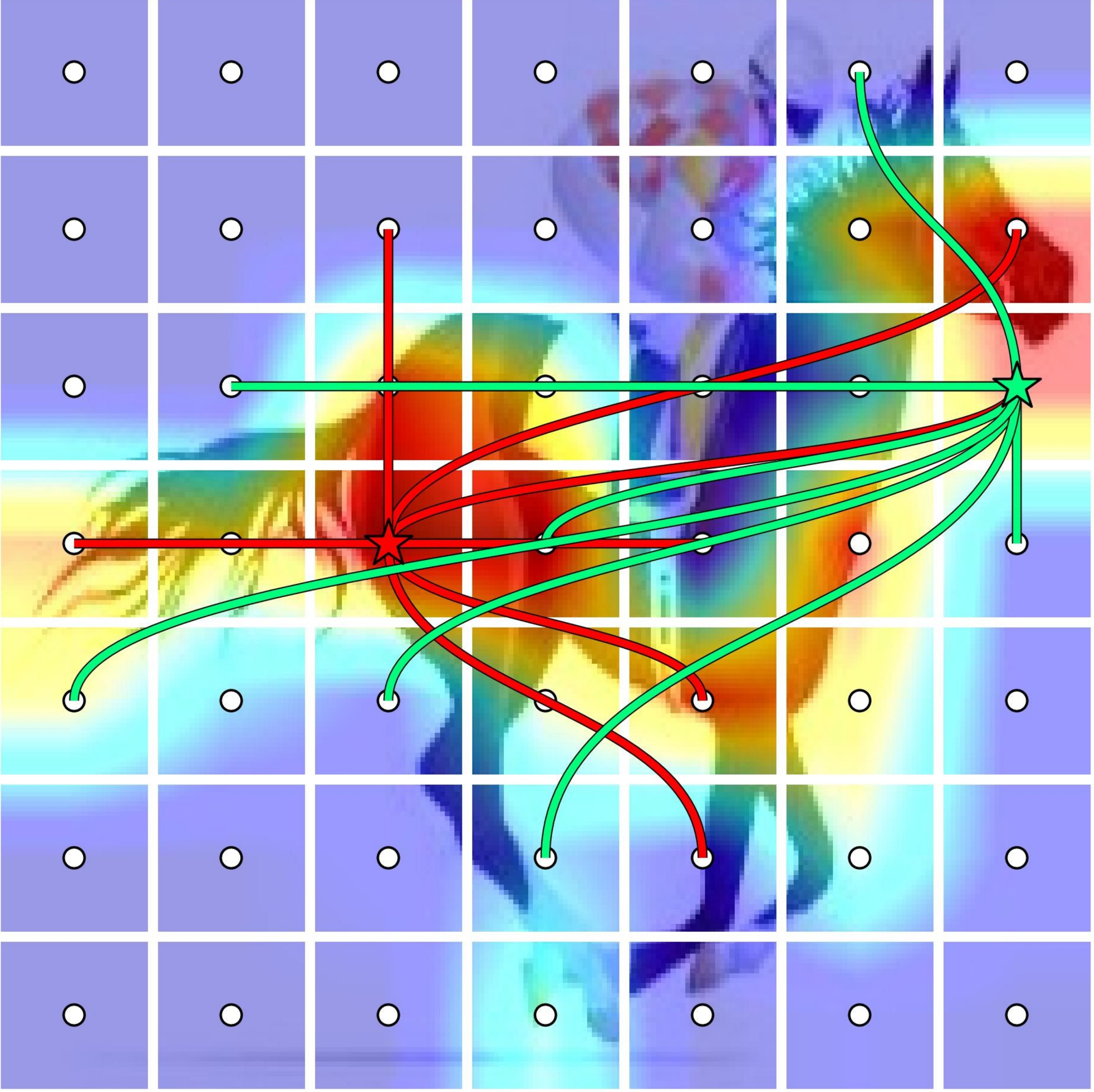}}} & \multicolumn{1}{c}{\frame{\includegraphics[width=0.3\linewidth]{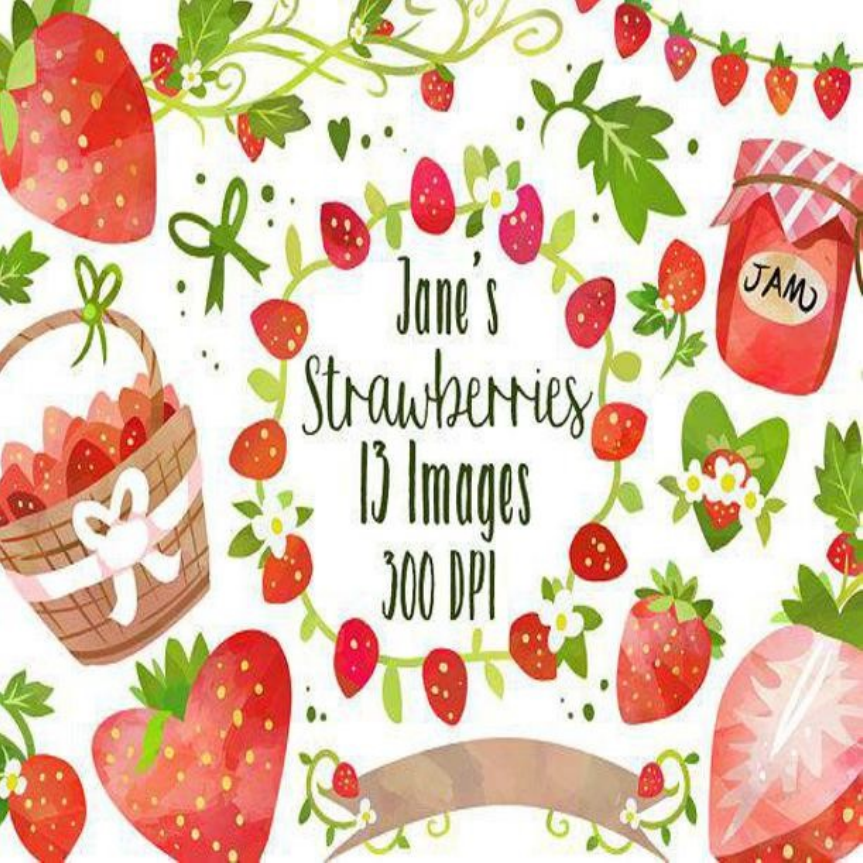}}} & \multicolumn{1}{c}{\frame{\includegraphics[width=0.3\linewidth]{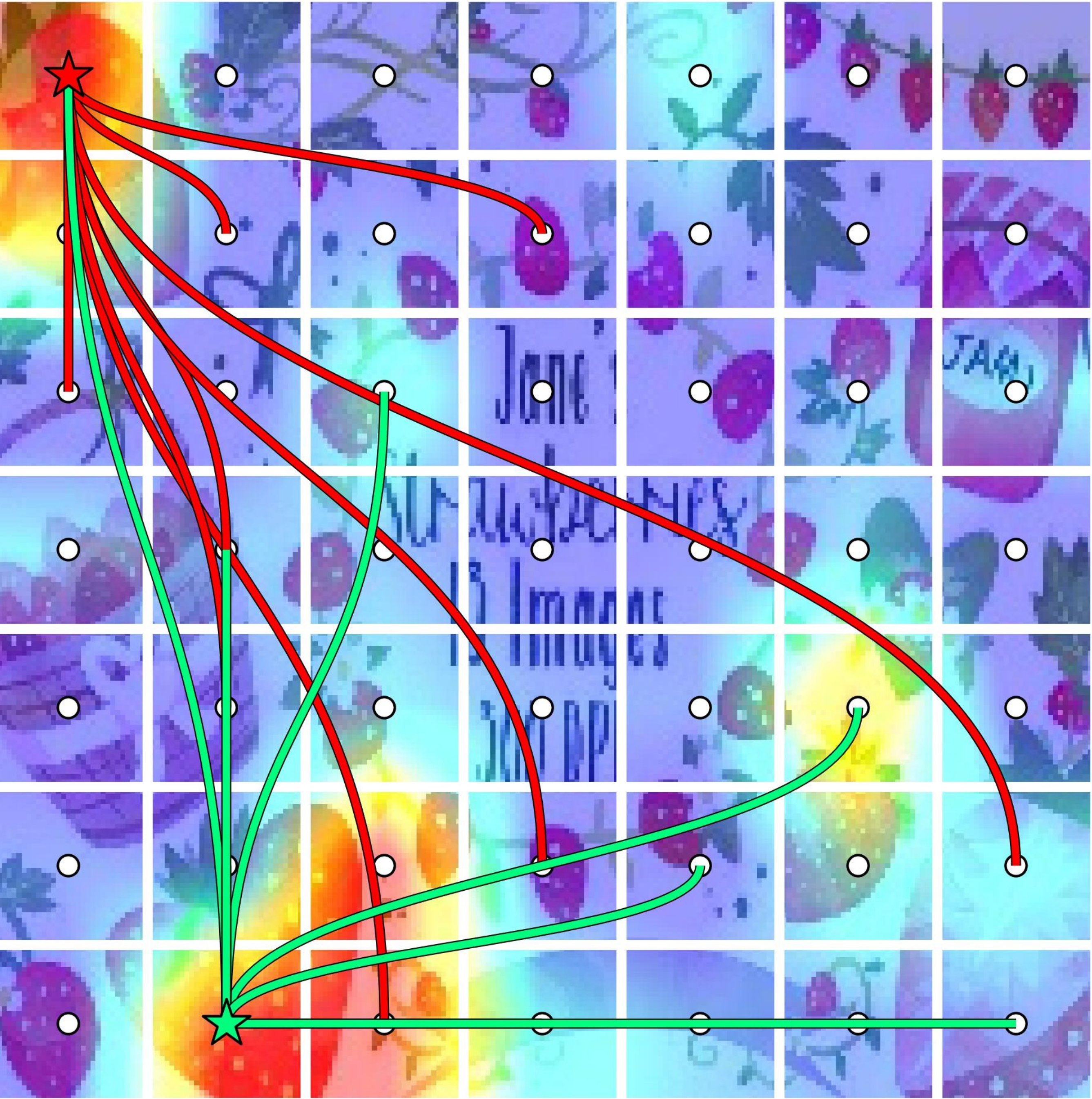}}} \\ \midrule
     \multicolumn{2}{c|}{(c) Single object robust to noise} & \multicolumn{2}{c}{(d) Multiple objects robust to noise} \\
      \bottomrule
    \end{tabular} 
    \end{adjustbox}
} 
\caption{GradCAM results extracted by the local graph. Please zoom in for viewing ease.}
\label{fig:gradcam} \vspace{-0.4cm}
\end{figure}

\subsection{Qualitative Results} We present qualitative results of HiGDA, including GradCAM visualizations with the constructed local graph $\mathcal{G}^{(L)}$ in Fig. \ref{fig:gradcam} and t-SNE comparisons in Fig. \ref{fig:tsne}, to provide visible and explainable insights of the model operation that further demonstrates the superiority of HiGDA.

\subsubsection{GradCAM with \textit{LoG}.} Among the local nodes in $\mathcal{V}^{(L)}$, representing patches, two anchor nodes are selected where the model focuses on (based on GradCAM). Edges to their neighbors are visualized, with anchor nodes shown as stars ($\textcolor{red}{\star}$  and $\textcolor{green}{\star}$). Under HiGDA, the local graph $\mathcal{G}^{(L)}$ effectively connects highly correlated patches to each other. For a single-object image in Fig. \ref{fig:gradcam}a, $\mathcal{G}^{(L)}$ easily covers the entire \textit{dolphin} starting from two anchor nodes in the middle of the body. In the multiple-object image in Fig. \ref{fig:gradcam}b, it links objects with similar patterns such as \textit{pineapples}. Furthermore, $\mathcal{G}^{(L)}$ demonstrates robustness to noise by disregarding non-essential objects in Fig. \ref{fig:gradcam}c, focusing on a \textit{horse} while excluding the rider. Even in complex backgrounds, as illustrated in Fig. \ref{fig:gradcam}d, it accurately links relevant objects such as \textit{strawberries}, while omitting irrelevant ones.

\subsubsection{t-SNE comparisons.} 
\begin{figure}[!h]
\centering
    \begin{adjustbox}{width=0.47\textwidth,center}
    \begin{tabular}{cc}
      {\includegraphics[width=0.49\linewidth]{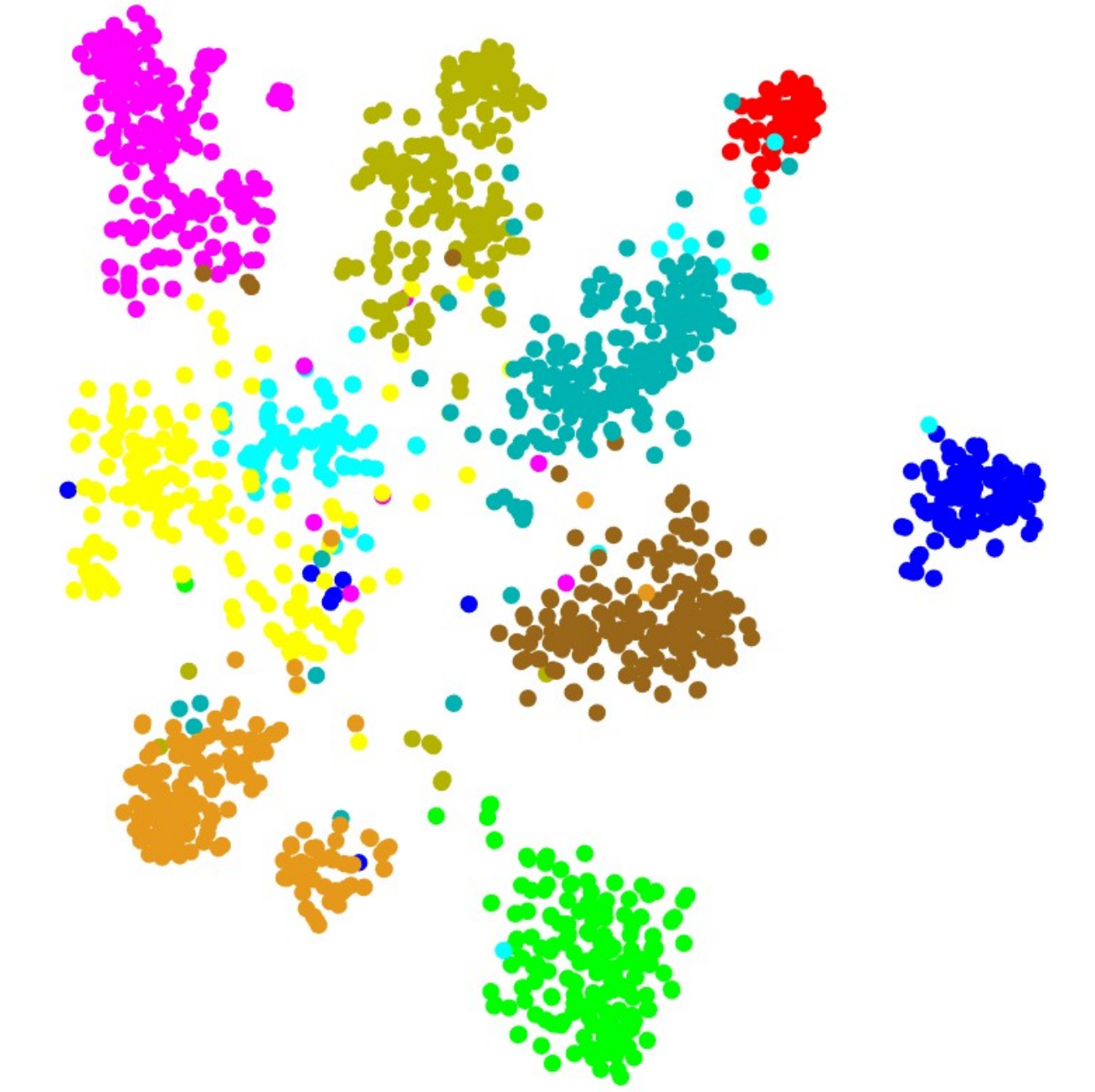}} &
      {\includegraphics[width=0.49\linewidth]{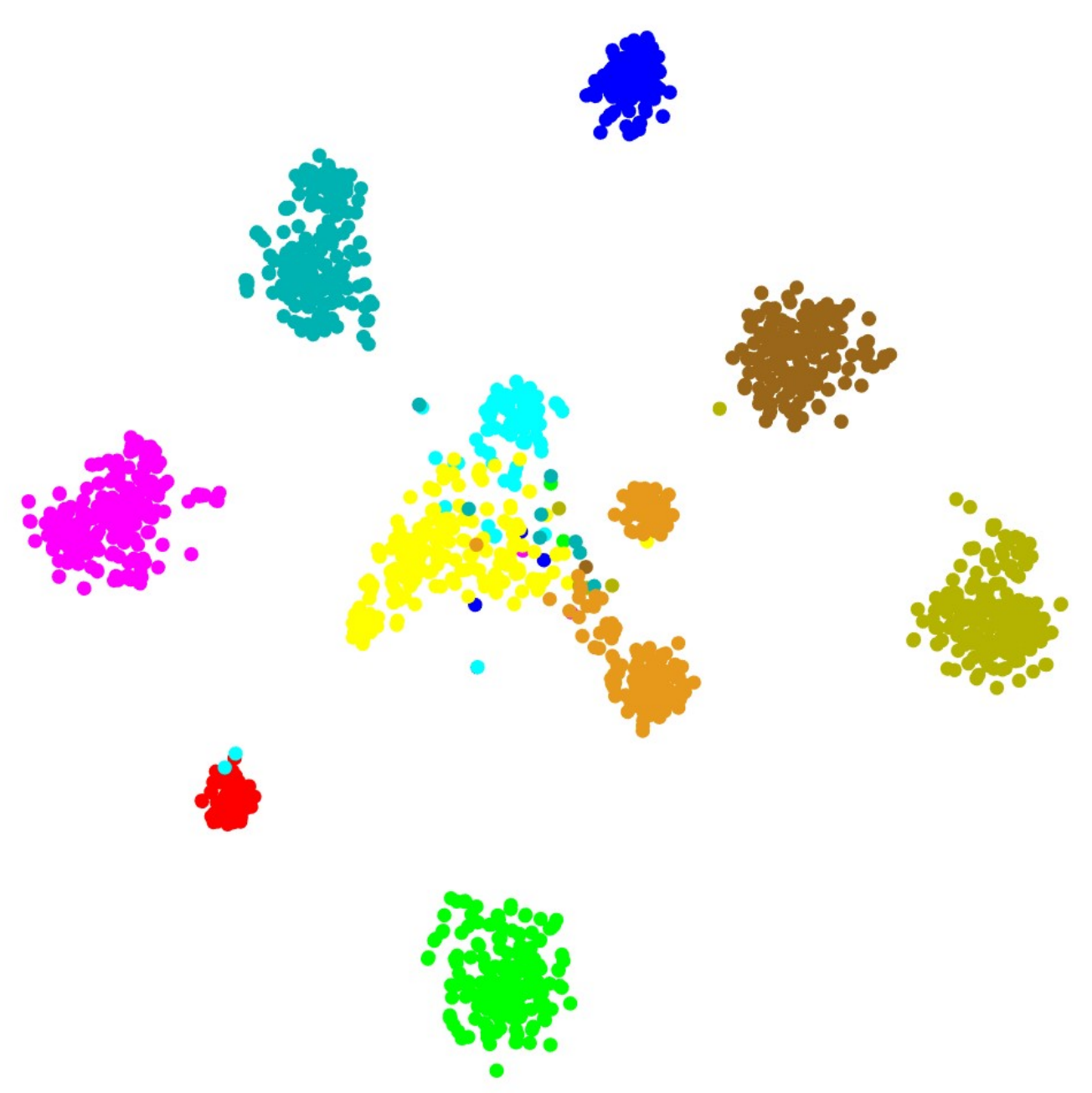}} \\
      (a) HiGDA-T & (b) HiGDA-T+GAL \\
      {\includegraphics[width=0.49\linewidth]{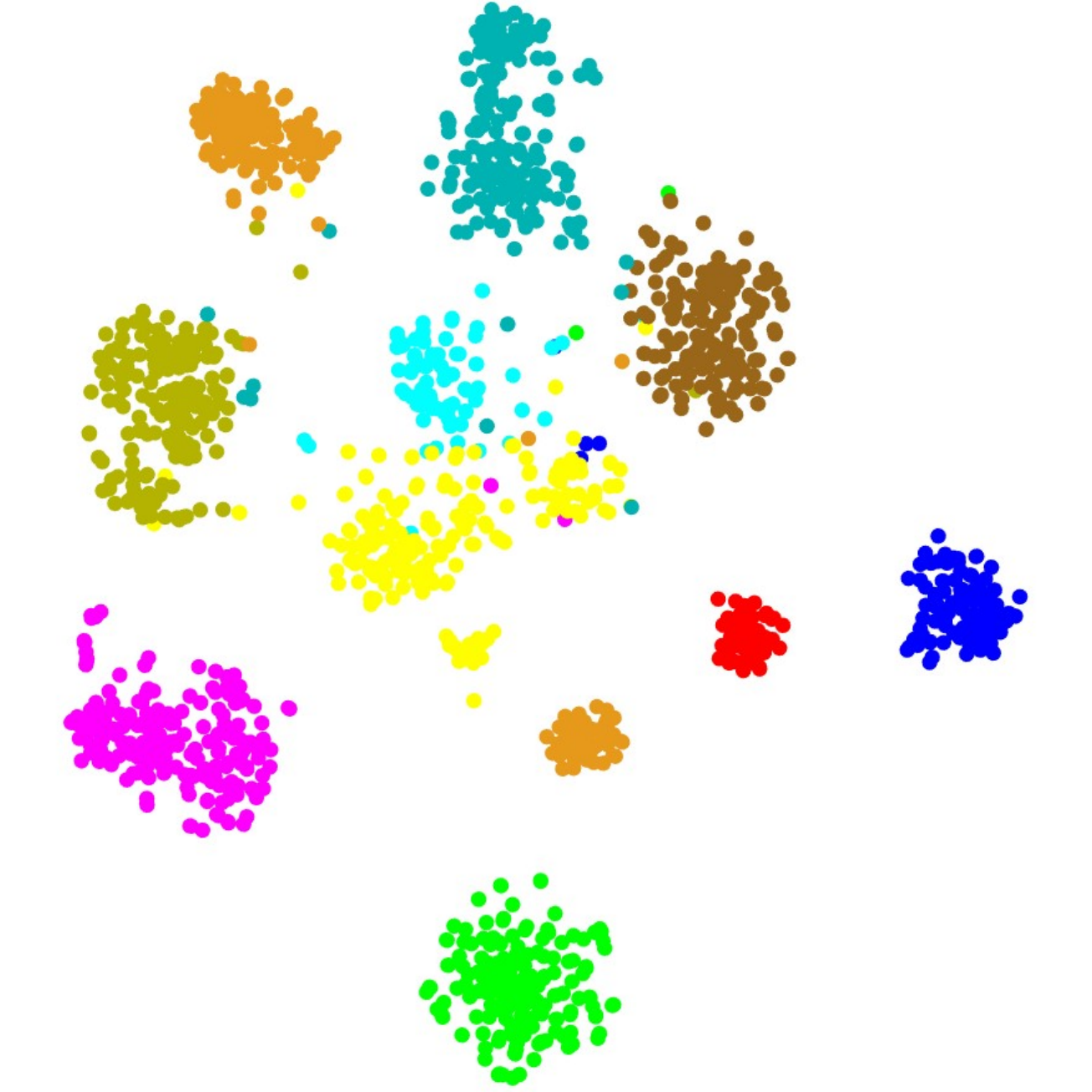}} &
      {\includegraphics[width=0.49\linewidth]{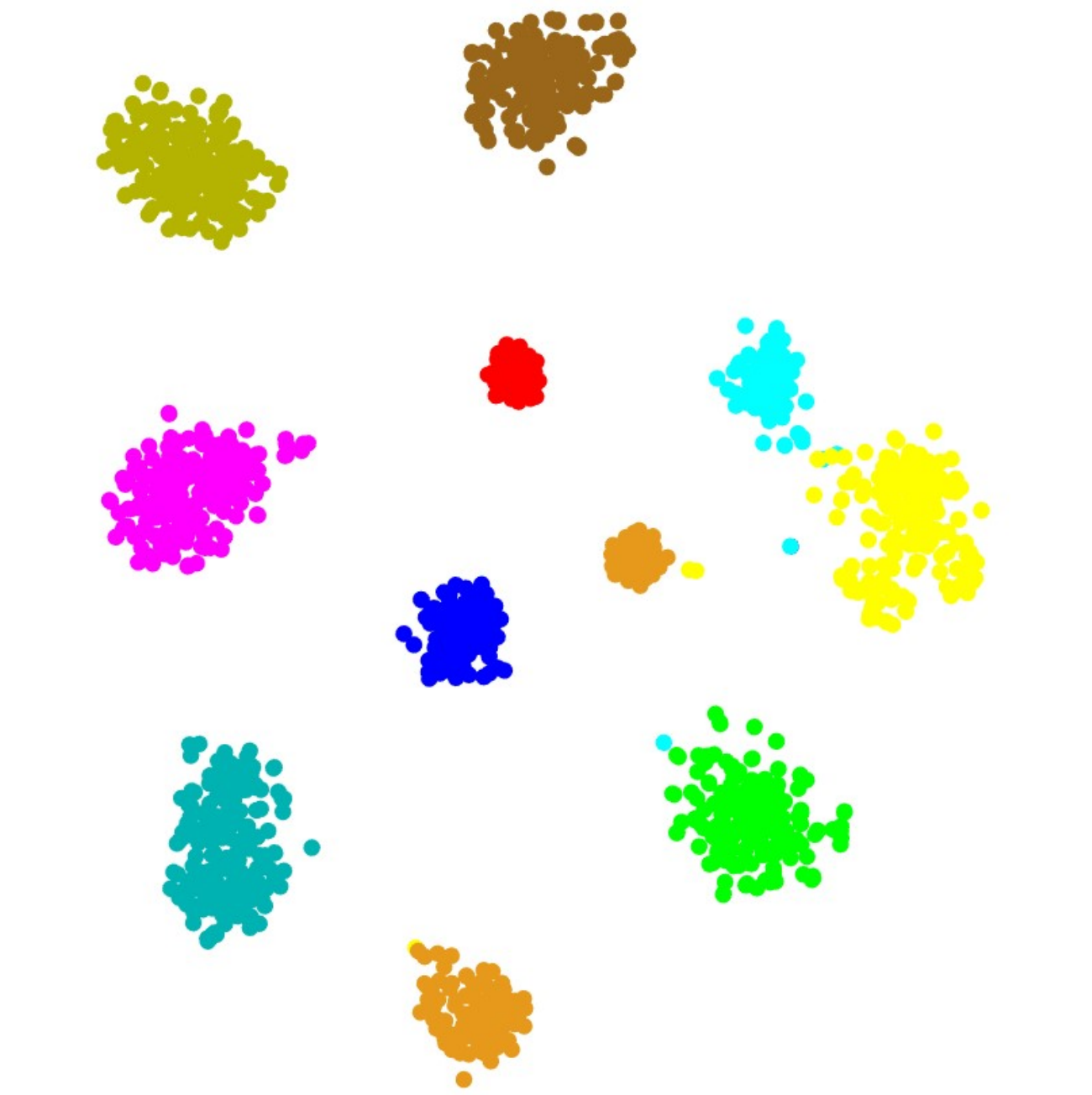}} \\
      (c) HiGDA-T+GAL+AAC & (d) HiGDA-T+GAL+MME \\ \\
      \multicolumn{2}{c}{{\includegraphics[width=1.05\linewidth]{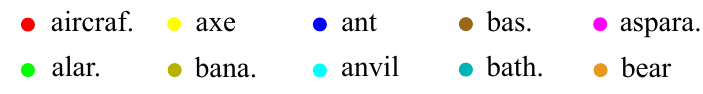}}}
    \end{tabular} 
    \end{adjustbox}
\caption{t-SNE \citep{tsne} visualization on \textit{DomainNet} of 10 classes in the \textit{real} to \textit{sketch} (rel$\rightarrow$skt) task. Please zoom in for viewing ease.
}
\label{fig:tsne} \vspace{-0.4cm}
\end{figure}

As illustrated in Fig. \ref{fig:tsne}a, the result indicates that the HiGDA-T baseline model with the S+T setting struggles to distinguish representations of different classes. In contrast, GAL proves to be effective by leading to better-distinguished representations in Fig. \ref{fig:tsne}b. However, the differences between samples of several categories, such as \textit{anvil}, \textit{axe}, and \textit{bear}, remain unclear. Consistent with the quantitative results, integrating HiGDA with MME produces higher quality representations than AAC, with samples from different categories well clustered, as shown in Fig. \ref{fig:tsne}c and Fig. \ref{fig:tsne}d, respectively. 

\vspace{-0.2cm}
\section{Conclusion}

In this study, we introduced a hierarchical graph of nodes to explicitly represent objects from the \textit{feature level} to the \textit{category level}. At the \textit{feature level}, an input image is divided into multiple patches, each conceptualized as a local node. Patches with strong relationships are connected to form a local graph, which better represents complex objects while filtering out irrelevant elements. At the \textit{category level}, each image is treated as a global node. We construct a global graph to aggregate the features of global nodes that share the same label information, thereby enriching the overall representations. Extensive experiments on various semi-supervised domain adaptation datasets, along with qualitative and quantitative analyses, demonstrate that the proposed method is more effective than previous approaches.
\subsection{Limitations} We found that $GoG$ is sensitive to noise, such as image samples that do not align well with the annotated labels, leading to cumulative errors and performance degradation. Controlling the contribution of node and edge loss to find the optimal solution is crucial to addressing this issue, which we plan to explore in future research.

\section{Acknowledgments}

This work was supported by the National Research Foundation of Korea (NRF) grant funded by the Korea Government (MSIT) (No. RS-2023-00214326 and No. RS-2023-00242528).

\bibliography{aaai25}
\end{document}